\documentclass[conference]{IEEEtran}
\IEEEoverridecommandlockouts
\usepackage{cite}
\usepackage{amsmath,amssymb,amsfonts}
\usepackage{algorithmic}
\usepackage{graphicx}
\usepackage{textcomp}
\usepackage{xcolor}
\def\BibTeX{{\rm B\kern-.05em{\sc i\kern-.025em b}\kern-.08em
    T\kern-.1667em\lower.7ex\hbox{E}\kern-.125emX}}

\usepackage{amsmath}
\usepackage{bm}

\usepackage[linesnumbered,ruled,vlined]{algorithm2e}
\SetKwInOut{Input}{Input}

\usepackage{booktabs}
\usepackage{multirow}

\usepackage{subfig}

\usepackage{cleveref}
\crefname{section}{\S}{\S\S}
\crefname{table}{Table}{}
\crefname{figure}{Fig.}{}
\crefname{algorithm}{Algorithm}{}
\crefname{equation}{}{}
\crefname{appendix}{Appendix}{}
\crefformat{section}{\S#2#1#3}  %

\newcommand{\De}{\ensuremath{\mathrm{\texttt{De}}}}
\newcommand{\En}{\ensuremath{\mathrm{\texttt{En}}}}
\newcommand{\Es}{\ensuremath{\mathrm{\texttt{Es}}}}
\newcommand{\Fr}{\ensuremath{\mathrm{\texttt{Fr}}}}
\newcommand{\Ar}{\ensuremath{\mathrm{\texttt{Ar}}}}
\newcommand{\Ru}{\ensuremath{\mathrm{\texttt{Ru}}}}
\newcommand{\Zh}{\ensuremath{\mathrm{\texttt{Zh}}}}

\begin{document}

\title{Improving Zero-shot Neural Machine Translation \\on Language-specific Encoders-Decoders
}

\author{\IEEEauthorblockN{Junwei Liao}
\IEEEauthorblockA{\textit{Department of Computer Science} \\
\textit{University of Electronic Science and Technology of China}\\
Chengdu, China \\
junwei.liao@outlook.com}
\and
\IEEEauthorblockN{Yu Shi}
\IEEEauthorblockA{\textit{Cognitive Services Research Group} \\
\textit{Microsoft}\\
Seattle, USA \\
yushi@microsoft.com}
\and
\IEEEauthorblockN{Ming Gong}
\IEEEauthorblockA{\textit{STCA NLP Group} \\
\textit{Microsoft}\\
Beijing, China \\
migon@microsoft.com}
\and
\IEEEauthorblockN{Linjun Shou}
\IEEEauthorblockA{\textit{STCA NLP Group} \\
\textit{Microsoft}\\
Beijing, China \\
lisho@microsoft.com}
\and
\IEEEauthorblockN{Hong Qu}
\IEEEauthorblockA{\textit{Department of Computer Science} \\
\textit{University of Electronic Science and Technology of China}\\
Chengdu, China \\
hongqu@uestc.edu.cn}
\and
\IEEEauthorblockN{Michael Zeng}
\IEEEauthorblockA{\textit{Cognitive Services Research Group} \\
\textit{Microsoft}\\
Seattle, USA \\
nzeng@microsoft.com}
}

\maketitle

\begin{abstract}
Recently, universal neural machine translation (NMT) with shared encoder-decoder gained good performance on zero-shot translation. Unlike universal NMT, jointly trained language-specific encoders-decoders aim to achieve universal representation across non-shared modules, each of which is for a language or language family. The non-shared architecture has the advantage of mitigating internal language competition, especially when the shared vocabulary and model parameters are restricted in their size. 
However, the performance of using multiple encoders and decoders on zero-shot translation still lags behind universal NMT.
In this work, we study zero-shot translation using language-specific encoders-decoders. We propose to generalize the non-shared architecture and universal NMT by differentiating the Transformer layers between language-specific and interlingua. By selectively sharing parameters and applying cross-attentions, we explore maximizing the representation universality and realizing the best alignment of language-agnostic information. We also introduce a denoising auto-encoding (DAE) objective to jointly train the model with the translation task in a multi-task manner.
Experiments on two public multilingual parallel datasets show that our proposed model achieves a competitive or better results than universal NMT and strong pivot baseline. Moreover, we experiment incrementally adding new language to the trained model by only updating the new model parameters. With this little effort, the zero-shot translation between this newly added language and existing languages achieves a comparable result with the model trained jointly from scratch on all languages.
\end{abstract}

\begin{IEEEkeywords}
multilingual neural machine translation, zero-shot, denoising auto-encoding, language-specific encoders-decoders
\end{IEEEkeywords}

\section{Introduction}

\begin{figure}[tb]
\centering
\subfloat[][Universal encoder-decoder]{
\includegraphics[width=0.47\columnwidth]{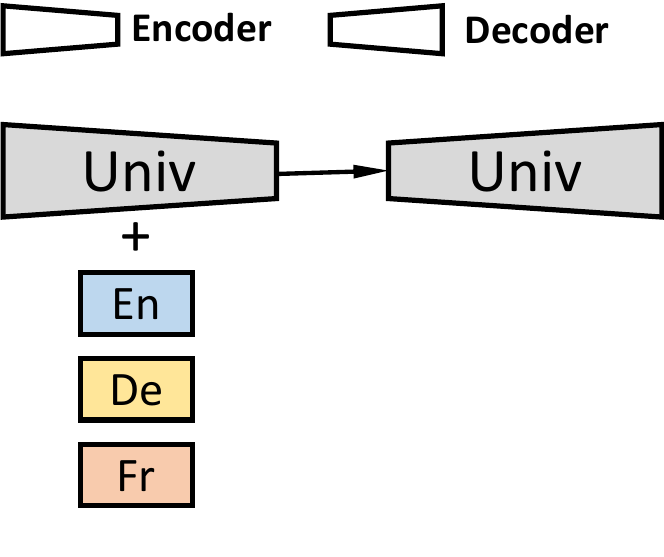}
\label{fig:universal}}
\subfloat[][Language-specific encoders-decoders]{
\includegraphics[width=0.47\columnwidth]{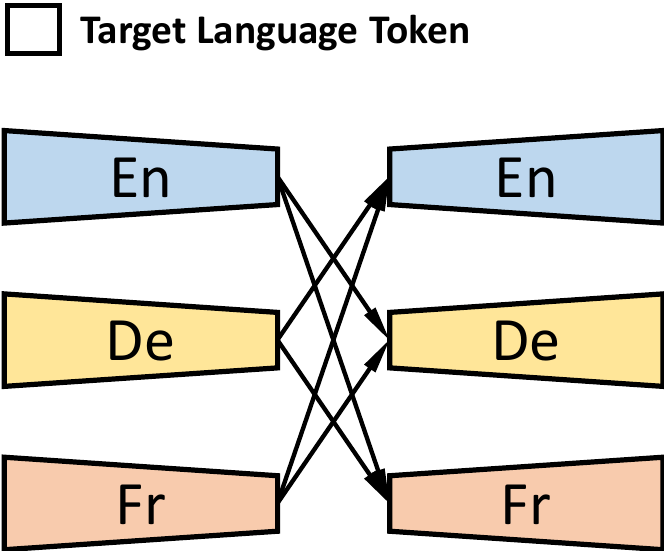}
\label{fig:language-specific}}
\caption{Two typical architectures of multilingual NMT.}
\label{fig1}
\end{figure}

Universal neural machine translation (NMT) draws much attention from the machine translation community in recent years, especially for zero-shot translation, which is first demonstrated by \cite{johnson2017google}. 
Later on, many subsequent works are proposed to further improve the zero-shot performance of universal NMT \cite{gu2019improved,ji2020cross,al2019consistency,arivazhagan2019missing,sestorain2018zero,he2016dual}. However, there are several shortcomings with this shared architecture (\cref{fig:universal}): (1) Since all languages share the same vocabulary and weights, the entire model needs to be retrained from scratch when adapting to a new group of languages; (2) The shared vocabulary grows dramatically when adding many languages, especially for those who do not share the alphabet, such as English and Chinese. The unnecessarily huge vocabulary makes the computation costly in decoding and hinders the deployment of commercial products; (3) This structure can only take text as input and cannot directly add raw signals from other modalities such as image or speech.

Another research direction in machine translation is the jointly trained language-specific encoders-decoders \cite{firat2016multi, lu2018neural, vazquez2019multilingual, escolano2019bilingual, escolano2020multilingual, escolano2020training}, as shown in \cref{fig:language-specific}. Due to the unshared architecture, these works' main goal is to improve the multilingual translation rather than zero-shot transfer. Hence the zero-shot performance of language-specific encoders-decoders still lags behind the universal model.
To solve this problem, previous works add extra network layers as ``interlingua" that is shared by all encoders and decoders \cite{firat2016multi, lu2018neural, vazquez2019multilingual}. 

In this paper, we focus on improving the zero-shot translation on language-specific encoders-decoders. We propose a novel interlingua mechanism that leverages the advantages of both architectures. Specifically, we fully exploit the characteristics of Transformer architecture \cite{vaswani2017attention} without introducing extra network layers. By selectively sharing parameters and applying cross-attentions, we explore maximizing the representation universality and realizing the best alignment of language-agnostic information. Our method can provide good enough universal representation for zero-shot translation, making explicit representation alignment not necessary.

Previous works have proved that monolingual data is helpful to zero-shot translation \cite{lu2018neural, vazquez2019multilingual, zhu2020language}, and reconstruction objective is usually used. However, we found that jointly train the model using a denoising auto-encoding (DAE) task together with the translation task is much more beneficial to zero-shot transfer. We argue that denoising objective can exploit monolingual data more efficiently than reconstruction objective which brings a spurious correlation between source and target sentences.

We verify our method on two public multilingual datasets. Europarl has four languages (En, De, Fr, Es) from close related languages. MultiUN has a group of distant languages (En, Ar, Ru, Zh). The results show that our approach achieve competitive or better results than pivot-based baseline and universal architecture on zero-shot translation. Furthermore, we show that our model can add new languages using incremental training without retraining the existing modules.

Our main contributions can be summarized as:
\begin{itemize}
\item We focus on improving zero-shot translation of language-specific encoders-decoders. It keeps the advantage of adding new languages without retraining the system and achieves comparable or better performance than universal encoder-decoder counterpart and pivot-based methods at the same time.
\item We propose a novel interlingua mechanism within the Transformer architecture without introducing extra network layers. We also propose a multi-task training of DAE and translation to exploit monolingual data more efficiently. These methods bring a significant improvement in the zero-shot performance of language-specific encoders-decoders.
\item We empirically explore several important aspects of proposed methods and give a detailed analysis.
\end{itemize}

\section{Related Works}

\subsection{Zero-shot Neural Machine Translation}
Zero-shot NMT has received more interest in recent years increasingly. \cite{gu2019improved} used decoder pretraining and back-translation to ignore spurious correlations in zero-shot translation; \cite{ji2020cross} proposed a cross-lingual pretraining on encoder before training the whole model with parallel data; \cite{al2019consistency} introduced a consistent agreement-based training method that encourages the model to produce equivalent translations of parallel sentences in auxiliary languages; \cite{arivazhagan2019missing} exploited an explicit alignment loss to align the sentence representations of different languages with same means in the high-level latent space; \cite{sestorain2018zero} proposed a fine-tuning technique that uses dual learning \cite{he2016dual}.
These works all use universal encoder-decoder, while our approaches adopt language-specific encoders-decoders.

\subsection{Language-specific Encoders-Decoders}
Most of the works on multilingual NMT use universal encoder-decoder.
Few works studied the language-specific encoders-decoders that is more flexible than universal encoder-decoder. 
\cite{firat2016multi} proposed extending the bilingual recurrent NMT architecture to the multilingual case by designing a shared attention-based mechanism between the language-specific encoders and decoders. 
\cite{lu2018neural} introduced a neural interlingua into language-specific encoders-decoders NMT architecture that captures language-independent semantic information in its sentence representation. \cite{vazquez2019multilingual} incorporated a self-attention layer, shared among all language pairs, that serves as a neural interlingua. 
A series of works \cite{escolano2019bilingual,escolano2020multilingual,escolano2020training} 
adopted different training strategies to improve multilingual NMT performance on language-specific encoders-decoders without any shared parameters. Although these works make some progress on improving the performance of multilingual translation at different levels, they still lag behind universal encoder-decoder on zero-shot translation.

\subsection{Parameter Shared Methods for Multilingual NMT}
Several proposals promote cross-lingual transfer, the key to zero-shot translation, by modifying the model’s architecture and selectively sharing parameters.
\cite{blackwood2018multilingual} proposed sharing all parameters but the attention mechanism. 
\cite{zaremoodi2018adaptive} utilized recurrent units with multiple blocks together with a trainable routing network.
\cite{platanios2018contextual} develop a contextual parameter generator that can be used to generate the encoder-decoder parameters for any source-target language pair. 
Our work uses parameter sharing based on Transformer \cite{vaswani2017attention}. The most similar work like ours is \cite{sachan2018parameter} but with some major differences:
(1)They study the parameter sharing methods only on one-to-many translation while we focus on zero-shot translation in many-to-many scenario; (2) They share the partial self-attention weights in all decoders layers while we share the selective layers of encoders; (3) They use one shared vocabulary while we use separated vocabulary for each language, which is more flexible to add new languages without retraining the whole system. 

\section{Approach}

In this section, we first introduce the language-specific encoders-decoders adopted in our method. Then we propose our approach -- combining interlingua via parameter sharing and DAE task jointly trained with translation task -- to improve the zero-shot translation on language-specific encoders-decoders.

\subsection{Background: Language-specific Encoders-Decoders for Multilingual NMT}

Language-specific encoder-decoder for multilingual NMT is based on the sequence-to-sequence model except that, as \cref{fig:language-specific} illustrates, each language has its own encoder and decoder. We denote the encoder and the decoder for the $i_{th}$ language in the system as $enc_i$ and $dec_i$, respectively. For language-specific scenarios, both the encoder and decoder are considered independent modules that can be freely interchanged to work in all translation directions. 
We use $(x_i, y_j)$ where $i, j \in \{1,...,K\}$ to represent a pair of sentences translating from a source language $i$ to a target language $j$. $K$ languages are considered in total. Our model is trained by maximizing the likelihood over training sets $D_{i,j}$ of all available language pairs $\mathcal{S}$. 
Formally, we aim to maximize $\mathcal{L}_{mt}$: 
\begin{equation}
    \mathcal{L}_{mt}(\theta) = 
    \sum_{\begin{subarray}{c}(x_i, y_j)\in D_{i, j},\\ (i, j) \in \mathcal{S}\end{subarray}}
    \log p(y_j | x_i; \theta),
    \label{eq1}
\end{equation}
where the probability $p(y_j | x_i)$ is modeled as
\begin{equation}
    p(y_j | x_i) = dec_j(enc_i(x_i)).
    \label{eq2}
\end{equation}

\cite{johnson2017google} showed that a trained multilingual NMT system could automatically translate between unseen pairs without any direct supervision, as long as both source and target languages were included in training. In other words, a model trained, for instance, on Spanish $\rightarrow$ English and English $\rightarrow$ French can directly translate from Spanish to French. Such an emergent property of a multilingual system is called zero-shot translation. It is conjectured that zero-shot NMT is possible because the optimization encourages different languages to be encoded into a shared space so that the decoder is detached from the source languages.
Universal encoder-decoder can naturally possess this property because all language pairs share the same encoder and decoder. But for language-specific encoders-decoders, there is no shared part among languages. That makes transfer learning hardly take effect to transfer the knowledge learned in high-source language to low-source language. That explains why language-specific encoders-decoders underperform universal encoder-decoder on zero-shot translation. \cite{escolano2020multilingual} also attributes this problem to limited shared information. To improve the zero-shot translation on language-specific encoders-decoders, we propose the interlingua via parameter sharing and DAE task.


\subsection{Interlingua via Parameter Sharing and Selective Cross-attention}
\begin{figure}[tb]
\centering
\includegraphics[width=1\columnwidth]{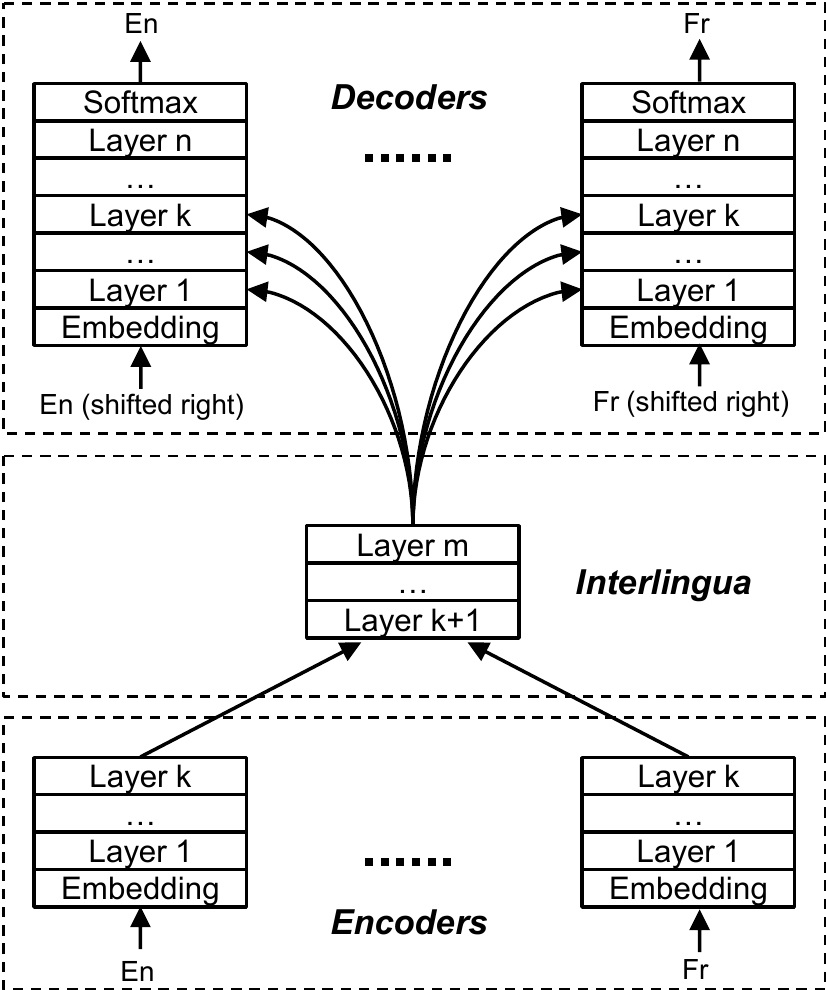} 
\caption{Interlingua via parameter sharing and selective cross-attention. We use English and French as two example languages for illustrating.}
\label{fig:interlingua}
\end{figure}

Previous works \cite{firat2016multi,lu2018neural,vazquez2019multilingual} propose to use an interlingua layer to bring the shared space to language-specific encodes-decoders. Interlingua is a shared component between encoders and decoders, which maps the output of language-specific encoders into a common space and gets an intermediate universal representation as the input to decoders. They implemented interlingua as extra network layers that make the model complicated and training inefficient. Unlike their methods, we propose implementing an interlingua by sharing the Transformer \cite{vaswani2017attention} layers of language-specific encoders. \cref{fig:interlingua} gives a schematic diagram for our proposed model.

\cite{conneau2020emerging,artetxe2020cross} show that transfer is possible even when there is no shared vocabulary across the monolingual corpora. The only requirement is that there are some shared parameters in the top layers of the multilingual encoder. Inspired by their work, we share the top layers of language-specific encoders while keeping the embedding and low layers intact. In this way, the model can get the transfer learning ability via parameter sharing while maintaining the multi-encoders' flexibility via independent vocabulary.
In the language-specific decoders' side, we keep the decoders separately without sharing parameters. 

Intuitively, to generate zero-shot language sentences from interlingua representation in a common shared space, cross-attention between interlingua and decoder layers should capture the high-level information, such as semantics, which is language-agnostic. \cite{tan2019study} shows that for multilingual NMT, the top layers of the decoder capture more language-specific information. Base on their conclusion, we can infer that low layers of decoder capture more language-agnostic information. Combining the two points above, we only conduct cross-attention between the interlingua and low layers of the decoder to capture language-agnostic knowledge more effectively.
We denote the $enc'_i$ as the language-specific part of encoder $i$, $dec'_j$ as the decoder $j$ with modified cross-attention. Equation \cref{eq2} is updated to
\begin{equation}
    p(y_j | x_i) = dec'_j(interlingua(enc'_i(x_i))),
    \label{eq3}
\end{equation}
where $interlingua$ denotes the interlingua via shared Transformer layers. 

\subsection{Denoising Auto-encoder (DAE) Task}
We propose to use a DAE task and train with translation task jointly.
Our training data cover $\mathcal{K}$ languages, and each $D_k$ is a collection of monolingual sentences in language $k$. We assume access to a noising function $g$, defined below, that corrupts text, and train the model to predict the original text $x_k$ given $g(x_k)$. More formally, we aim to maximize $\mathcal{L}_{dae}$ as
\begin{equation}
    \mathcal{L}_{dae}(\theta) = 
    \sum_{\begin{subarray}{c}x_k \in D_k,\\ k \in \mathcal{K}\end{subarray}}
    \log p(x_k | g(x_k); \theta),
    \label{eq4}
\end{equation}
where $x_k$ is an instance in language $k$ and the probability $p(x_k | g(x_k))$ is defined by
\begin{equation}
    p(x_k | g(x_k)) = dec'_k(interlingua(enc'_k(g(x_k)))).
    \label{eq5}
\end{equation}

The noising function $g$ injects three types of noise to obtain the randomly perturbed text: First, we randomly drop tokens of the sentence with a probability; Second, we substitute tokens with the special masking token with another probability; Third, the token order is locally shuffled. The candidate tokens can be in a unit of subword or whole word and span to N-grams.

Finally, the objective function of our learning algorithm is
\begin{equation}
    \mathcal{L} = \mathcal{L}_{mt} + \mathcal{L}_{dae}. 
    \label{eq8}
\end{equation}
We jointly train the multiple encoders and decoders by 
randomly selecting the two tasks with equal probability, i.e. multilingual translation and DAE task.

\section{Experiments}
\subsection{Setup}
We evaluate the proposed approaches against several strong baselines on two public multilingual datasets across a variety of language, Europarl\footnote{http://www.statmt.org/europarl/} \cite{koehn2005europarl} and MultiUN\footnote{http://opus.nlpl.eu/MultiUN.php} \cite{eisele2010multiun}. In all experiments, we use BLEU \cite{papineni2002bleu} as the automatic metric for translation evaluation.\footnote{We calculate BLEU scores using toolkit SacreBLEU \cite{post2018call} with default tokenizer except Chinese which uses Zh tokenizer.}

\subsubsection{Datasets}

\begin{table}[tb]
	\caption{Overall dataset statistics where each pair has a similar number of training samples. All the remaining directions are used to evaluate zero-shot performance. 
	}
	\centering
	\scalebox{0.9}{
	\begin{tabular}{l | l r }
	\toprule
	Dataset & parallel pairs & size/pair \\
	\midrule
    Europarl   & Es-En, De-En, Fr-En  & 0.6M \\
    \midrule
    MultiUN & Ar-En, Ru-En, Zh-En & 2M \\
    \bottomrule

	\end{tabular}}
	\label{table:data}
	\vspace{0pt}
\end{table}

The detailed statistics of Europarl and MultiUN datasets are in \cref{table:data}. To simulate the zero-shot settings, the training set only allowing parallel sentences from/to English, where English acts as the pivot language. For Europarl corpus, to compare with previous work fairly, we follow \cite{al2019consistency}'s method to preprocess the Europarl to avoid multi-parallel sentences in training data and ensure the zero-shot setting.
We use the \textit{dev2006} as the validation set and the \textit{test2006} as the test set which contain 2,000 multi-parallel sentences. For vocabulary, We use SentencePiece \cite{kudo2018sentencepiece} to encode text as WordPiece tokens \cite{kudo2018subword}. Due to the language-specific architecture of our model, each language has its own vocabulary. We choose a vocabulary of 32K wordpieces for each language. 

For MultiUN corpus, 
We randomly sub-sample the 2M sentences per language pair for the training set and 2000 sentences per language pair for validation and test sets. Vocabulary size for each language is also 32K.

For DAE task, we only use monolingual data extracted from parallel training data. No extra monolingual data are used.

\subsubsection{Model}
Our code was implemented using PyTorch on top of the Huggingface Transformers library\footnote{https://github.com/huggingface/Transformers}. 
To decouple the output representation of an encoder from the task, we use the target language 
code as the initial token of the decoder input following the practice of \cite{gu2019improved}.
The Transformer layers in our model use the parameters of $d_{model}=512, d_{ff}=2048, n_{heads}=8$. The encoders have 9 layers with 6 top layers sharing parameters as interlingua. The decoders have 12 layers with 6 low layers conducting cross-attention with interlingua.
The output softmax layer is tied with input embeddings \cite{press2017using}.

\subsubsection{Training}
We train our models with the RAdam optimizer \cite{liu2019variance} and the inverse square root learning rate scheduler of \cite{vaswani2017attention} with 5e-4 learning rate and 64K linear warmup steps. For each model, we train it using 8 NVIDIA V100 GPUs with 32GB of memory and mixed precision. It takes around three days to train one model. We use batch size 32 for each GPU. We stop training at optimization step 200K and select the best model based on the validation set.
We search the following hyperparameter for zero-shot translation: batch size \{32, 96\}; learning rate \{5e-4, 1e-3, 2e-3\}; linear warmup steps \{4K, 10K, 32K, 40K, 64K\}.
We use dropout 0.1 throughout the whole model.
For decoding, we use beam-search with beam size 1 for all directions.
The noise function uses token deletion rate 0.2, token masking rate 0.1, WordPiece token as the token unit, and unigram as span width.

\subsubsection{Evaluation}
We focus our evaluation mainly on zero-shot performance of the following methods:
\begin{itemize}
    \item \textbf{Univ.}, which stands for directly evaluating a multilingual universal encoder-decoder model after standard training \cite{johnson2017google}.
    \item \textbf{Pivot}, which performs pivot-based translation using a multilingual universal encoder-decoder model (after standard training); often regarded as gold-standard.
    \item \textbf{Ours}, which represents language-specific encoders-decoders with shared interlingua and jointly trained with multilingual translation and DAE tasks.
\end{itemize}

To ensure a fair comparison in terms of model capacity, all the techniques above use the same 
Transformer architecture described above, i.e. 9 layers of encoder and 12 layers of decoder. All other results provided in the tables are as reported in the literature.

\subsection{Main Results}
The supervised and zero-shot translation results on Europarl and MultiUN corpora are shown in \cref{tab:europarl} and \cref{tab:multiun}. 

\begin{table}[tb]
\caption{Zero-shot results on Europarl corpus.
}
\centering
\scriptsize
\def\arraystretch{0.8}
\setlength{\tabcolsep}{8.25pt}
\begin{tabular}{@{}lrr|rr|r@{}}
\toprule
                        & \multicolumn{2}{c|}{Previous work}   & \multicolumn{2}{c|}{Our baselines} \\
\cmidrule[0.5pt]{2-5}
                        & BRLM-SA$^\dagger$       & Agree$^\ddagger$     & Univ. & Pivot     & Ours \\
\cmidrule[0.5pt]{1-6}                                                                        
$\En \rightarrow \De$   & 24.95            & 22.44            & 26.05 & ---             & \textbf{26.21} \\
$\De \rightarrow \En$   & ---              & 29.07            & \textbf{31.83} & ---    & 31.81 \\
$\En \rightarrow \Fr$   & \textbf{35.91}   & 32.55            & 34.40 & ---             & 34.29 \\
$\Fr \rightarrow \En$   & ---              & 33.30            & 35.02 & ---             & \textbf{35.40} \\
$\En \rightarrow \Es$   & 34.92            & 33.80            & \textbf{35.46} & ---    & 34.98 \\
$\Es \rightarrow \En$   & ---              & 34.53            & 36.34 & ---             & \textbf{36.37} \\
\cmidrule[0.5pt]{1-6}                                                                   
Supervised (avg.)       & ---              & 30.95            & 33.18 & ---             & \textbf{33.18} \\
\cmidrule[0.5pt]{1-6}                                                                   
$\De \rightarrow \Fr$   & \textbf{30.66}   & 24.45            & 27.35 & 29.35           & 29.41 \\
$\Fr \rightarrow \De$   & ---              & 19.15            & 20.05 & 25.02           & \textbf{25.28} \\
$\De \rightarrow \Es$   & ---              & 22.45            & 28.28 & 30.10           & \textbf{30.48} \\
$\Es \rightarrow \De$   & ---              & 20.70            & 20.55 & 25.15           & \textbf{25.50} \\
$\Fr \rightarrow \Es$   & 37.02            & 29.91            & 33.50 & 34.75           & \textbf{38.21} \\
$\Es \rightarrow \Fr$   & ---              & 30.94            & 34.24 & 34.90           & \textbf{38.20} \\
\cmidrule[0.5pt]{1-6}                                                                   
Zero-shot (avg.)        & ---              & 24.60            & 27.33 & 29.88           & \textbf{31.18} \\
\bottomrule
\end{tabular}\\
$^\dagger$BRLM-SA \cite{ji2020cross}. $^\ddagger$Agree \cite{al2019consistency}.
\label{tab:europarl}
\end{table}

\begin{table}[tb]
\caption{Zero-shot results on MultiUN corpus.
}
\centering
\scriptsize
\def\arraystretch{0.8}
\setlength{\tabcolsep}{8.25pt}
\begin{tabular}{@{}lr|rr|r@{}}
\toprule
                        & Previous work         & \multicolumn{2}{c|}{Our baselines}  \\
\cmidrule[0.5pt]{2-4}
                        & ZS+LM$^\dagger$       & Univ. & Pivot               & Ours  \\
\cmidrule[0.5pt]{1-5}
$\En \rightarrow \Ar$   & ---                     & 37.07 & ---               & \textbf{38.22} \\
$\Ar \rightarrow \En$   & ---                     & 51.82 & ---               & \textbf{52.82} \\
$\En \rightarrow \Ru$   & ---                     & 40.83 & ---               & \textbf{42.02} \\
$\Ru \rightarrow \En$   & ---                     & 47.52 & ---               & \textbf{48.57} \\
$\En \rightarrow \Zh$   & ---                     & 52.51 & ---               & \textbf{52.64} \\
$\Zh \rightarrow \En$   & ---                     & 47.74 & ---               & \textbf{48.63} \\
\cmidrule[0.5pt]{1-5}                                                   
Supervised (avg.)       & 45.80                   & 46.25 & ---               & \textbf{47.15} \\
\cmidrule[0.5pt]{1-5}                                                           
$\Ar \rightarrow \Ru$   & 21.5                    & 27.96 & \textbf{36.68}    & 35.04 \\
$\Ru \rightarrow \Ar$   & 28.0                    & 25.65 & \textbf{30.95}    & 29.33 \\
$\Ar \rightarrow \Zh$   & \textbf{43.8}           & 37.96 & 43.64             & 41.18 \\
$\Zh \rightarrow \Ar$   & 27.3                    & 22.67 & \textbf{28.46}    & 26.27 \\
$\Ru \rightarrow \Zh$   & \textbf{43.3}           & 31.70 & 36.33             & 34.38 \\
$\Zh \rightarrow \Ru$   & 19.9                    & 21.86 & \textbf{29.05}    & 25.56 \\
\cmidrule[0.5pt]{1-5}                                                  
Zero-shot (avg.)        & 30.6                    & 27.97 & \textbf{34.18}    & 31.96 \\
\bottomrule
\end{tabular}\\
$^\dagger$ZS+LM \cite{gu2019improved}. 
\label{tab:multiun}
\end{table}

\subsubsection{Results on Europarl Dataset} 
For zero-shot translation, our model outperforms all other baselines except on De$\rightarrow$Fr to BRSM-SA \cite{ji2020cross} that use a large scale monolingual data to pretrain encoder. Especially, Our model outperforms the pivot-based translation in all directions. For Fr$\leftrightarrow$Es pairs, our model improves about 3.4 BLEU points over the pivoting. The pivot-based translation is a strong baseline in the zero-shot scenario that often beats the other multilingual NMT system baselines (\cite{johnson2017google,al2019consistency}). 
Pivoting translates source to pivot then to target in two steps, causing an inefficient translation process. Our approaches translate source to target directly between any zero-shot directions, which is more efficient than pivoting. 
For supervised translation, our model gets the same averaged score with universal NMT and receives a better score in most direction (En$\rightarrow$De, Fr$\rightarrow$En, Es$\rightarrow$En). Baseline Agree \cite{al2019consistency} using agreement loss manifests the worst performance. We conjecture that it is because their model is based on LSTM \cite{hochreiter1997long}, which was proved inferior to Transformer on multilingual NMT\cite{lakew2018comparison}. But regarding the BRLM-SA \cite{ji2020cross} that uses Transfomer-big and cross-lingual pretraining with large scale monolingual, our model only uses the parallel data and get a better score on most of supervised and zero-shot direction.

\subsubsection{Results on MultiUN Dataset}
Our model performs better for supervised and zero-shot translation than universal NMT and the enhanced universal NMT with language model pretraining for decoder \cite{gu2019improved} that denoted as ZS+LM in \cref{tab:multiun}. But our model still lags behind the pivoting 
about 2.2 BLEU points. Similar results were observed in previous works \cite{gu2019improved,ji2020cross} where their methods based on universal encoder-decoder also underperform pivoting on MultiUN dataset. 
To surpass the pivoting, they introduced the back-translation \cite{sennrich2016improving} to generate pseudo parallel sentences for all zero-shot direction based on pretrained models, and further trained their model with these pseudo data. 
Strictly speaking, their methods are ``zero-resource'' translation rather than ``zero-shot'' translation since the model has seen explicit examples of zero-shot language pair during training, which zero-shot translation should not do \cite{johnson2017google}. 
Considering this reason, although our model can also benefit from adding training data augmented by back-translation, we decide not to use it to further improve our model just for beating the pivoting baseline.

\section{Discussion}
In this section, we discuss some important aspects of the proposed approach. 

\subsection{Incremental Language Expansion}

First, we explore our model's zero-shot transfer capability when incrementally adding new languages. The incremental training is similar to that in \cite{escolano2019bilingual}. 
For illustration, let us assume we have gotten a language-specific encoders-decoders model with initial training on En$\leftrightarrow$Es,Fr parallel data. Now we want to add a new language, say De, into the existing modules via incremental training only on En$\leftrightarrow$De parallel data. 
In the initial training, we jointly train three encoders and three decoders for languages En, Es, and Fr, respectively. 
In the incremental training step, we add both a new encoder and a new decoder for De into the initial model, share the interlingua layers with the initial modules, and randomly initialize the non-shared layers. The weight update is only conducted on the newly added non-shared layers using the En$\leftrightarrow$De parallel data by freezing the weights of the six initial modules. Both steps use the proposed multi-task training of DAE and translation objectives.

Following the same process, we also experimented with adding Fr or Es to the model initially trained on other three languages, respectively. 
We use the same experimental setting as in the main experiment unless stated otherwise. We compare the zero-shot translation results of the incremental training with the joint training on the Europarl dataset (En$\leftrightarrow$De,Fr,ES). 
The results are shown in \cref{tab:add-new-language}.

\begin{table}[tb]
\caption{The results of zero-shot translation on Europarl datasets when adding a new language. The metric is BLEU. \textbf{Joint}, \textbf{Init.}, \textbf{Incr.} represent joint training, initial training, and incremental training, respectively.} 
\centering
\footnotesize
\resizebox{1\linewidth}{!}{
\begin{tabular}[b]{l cc cc cc cc}
\toprule
Europarl & \multicolumn{8}{c}{De, Es, Fr $\leftrightarrow$ En} \\
\midrule
\multirow{2}{*}{Direction} & \multicolumn{2}{c}{Fr-De} & \multicolumn{2}{c}{Es-De} & \multicolumn{2}{c}{Es-Fr} & Zero & Parallel \\
& $\leftarrow$ & $\rightarrow$ & $\leftarrow$ & $\rightarrow$ & $\leftarrow$ & $\rightarrow$ & Avg & Avg \\
\midrule
\textbf{Joint} & 29.41 & 25.28 & 30.48 & 25.50 & 38.21 & 38.20 & \textbf{31.18} & \textbf{33.18} \\
\midrule
\multicolumn{9}{l}{\textit{Intial training on En$\leftrightarrow$Es,Fr, then incremental training on En$\leftrightarrow$De.}} \\
\midrule
\textbf{Init.} &   --- &   --- &   --- &   --- & 37.99 & 37.40 & ---   & ---   \\
\textbf{Incr.} & 28.95 & 24.86 & 30.19 & 25.67 & 37.99 & 37.40 & 30.84 & 32.98 \\
\midrule
\multicolumn{9}{l}{\textit{Intial training on En$\leftrightarrow$De,Es, then incremental training on En$\leftrightarrow$Fr.}} \\
\midrule
\textbf{Init.} &   --- &   --- & 30.58 & 25.18 &   --- &   --- &   --- &   --- \\
\textbf{Incr.} & 29.61 & 24.58 & 30.58 & 25.18 & 37.57 & 37.84 & 30.89 & 32.78 \\
\midrule
\multicolumn{9}{l}{\textit{Intial training on En$\leftrightarrow$De,Fr, then incremental training on  En$\leftrightarrow$Es.}} \\
\midrule
\textbf{Init.} & 29.30 & 24.93 &   --- &   --- &   --- &   --- &   --- &   --- \\
\textbf{Incr.} & 29.30 & 24.93 & 30.60 & 24.82 & 38.24 & 38.03 & 30.99 & 32.83 \\
\bottomrule
\end{tabular}
}
\label{tab:add-new-language}
\end{table}

\cref{tab:add-new-language} shows that the universal representation learned in the initial model is easily transferred to the new language. The gap between the incremental training and the joint training is within 0.2$\sim$0.4 points on supervised translation and 0.27$\sim$0.34 points on zero-shot translation. We suspect that the hyperparameters for joint training may not be the best for incremental training. In addition, incremental training followed by continued joint training could further improve the performance. We leave those experiments in future work. 

Besides the good zero-shot transferability, incremental training is also lightweight. Due to the interlingua mechanism, we only update the model weights of the non-interlingua layers of the newly added modules. This dramatically reduces the number of trainable parameters. In contrast with the joint training that takes three days, incremental training only takes half a day to get a comparable result.

\subsubsection{Conclusion}
The proposed incremental training approach is a lightweight language expansion method that can fully leverage the initial model's interlingua layers and the universal representation learned from the initial training for both parallel and zero-shot languages.

\subsection{Interlingua Structure Analysis}

In this section, we empirically examine the following questions: 1) How important it is to share the interlingua layers of the encoders? 2) How to partition the language-specific and interlingua layers in the encoders given a fixed total number of layers? 3) Which layers in the decoders should be chosen to conduct cross-attention to the encoder output? 4) Whether sharing interlingua layers among the decoders can benefit the zero-shot translation?

Before the discussion, we first define some shorthands. We denote the encoder interlingua layers between layer $i$ and layer $j$ as \textbf{E$i$-$j$}, the decoder interlingua layers as \textbf{D$i$-$j$}, and the decoder layers that have cross-attention to interlingua output as \textbf{C$i$-$j$}. Both $i$ and $j$ are inclusive and 1-based.

We use the original Europarl corpus without preprocessing. 
To accelerate the experiment, we use a small version of the proposed model. It has a similar architecture to that in our main experiment except that the language-specific encoders have 8 layers and language-specific decoders have 10 layers. 
All the experiments use the same random seeds to ensure the results are comparable.

We design four groups of experiments to answer the questions. Because there can be exponentially many combinations considering all the different feasible sets of configuration, we intuitively only select a subset of these combinations.
The results are reported in \cref{tab:interlingua}.

\begin{table}[tb]
\caption{Dissecting the interlingua structure based on zero-shot performance. The best score in each group is bold. \textbf{E}/\textbf{D}/\textbf{C} represent encoder/decoder/cross-attention. The number after the letter denotes the range of shared encoder layers/shared decoder layers/cross-attention layers respectively.} 
\centering
\footnotesize
\resizebox{1\linewidth}{!}{
\begin{tabular}[b]{l cc cc cc cc}
\toprule
Europarl & \multicolumn{8}{c}{De, Es, Fr $\leftrightarrow$ En} \\
\midrule
\multirow{2}{*}{Direction} & \multicolumn{2}{c}{Fr-De} & \multicolumn{2}{c}{Es-De} & \multicolumn{2}{c}{Es-Fr} & Zero & Parallel \\
& $\leftarrow$ & $\rightarrow$ & $\leftarrow$ & $\rightarrow$ & $\leftarrow$ & $\rightarrow$ & Avg & Avg \\
\midrule
\multicolumn{9}{l}{\textit{1. Ablation of sharing encoders layers.}} \\
\midrule
\textbf{E3-8, C1-10} & 32.1 & 26.8 & 30.8 & 25.3 & 36.8 & 37.4 & \textbf{31.5} & \textbf{31.9} \\
\textbf{C1-10} & 15.5 & 8.6 & 12.4 & 9.3 & 15.1 & 19.1 & 13.3 & 31.6 \\
\midrule
\multicolumn{9}{l}{\textit{2. Change number of encoders shared layers.}} \\
\midrule
\textbf{E1-8, DC3-8} & 29.3 & 24.2 & 30.1 & 24.6 & 36.2 & 35.9 & \textbf{30.1} & 31.4 \\
\textbf{E3-8, DC3-8} & 30.1 & 23.9 & 30.6 & 25.2 & 35.0 & 34.5 & 29.9 & 31.5 \\
\textbf{E5-8, DC3-8} & 28.0 & 23.8 & 29.8 & 23.2 & 34.9 & 33.8 & 28.9 & 31.2 \\
\textbf{E7-8, DC3-8} & 27.3 & 22.1 & 28.2 & 21.3 & 33.7 & 31.4 & 27.3 & 31.7 \\
\textbf{DC3-8}       & 25.1 & 20.6 & 27.8 & 20.8 & 32.1 & 30.9 & 26.2 & \textbf{31.8} \\
\midrule
\multicolumn{9}{l}{\textit{3. Change number of decoder cross-attention layers.}} \\
\midrule
\textbf{E3-8, C5-10} & 31.3 & 26.0 & 29.6 & 24.6 & 36.7 & 37.4 & 30.9 & \textbf{32.0} \\
\textbf{E3-8, C3-8}  & 31.4 & 27.6 & 31.7 & 25.7 & 35.4 & 36.9 & 31.5 & 31.7 \\
\textbf{E3-8, C1-6}  & 31.6 & 27.3 & 32.1 & 26.4 & 37.5 & 37.9 & \textbf{32.1} & 31.8 \\
\midrule
\multicolumn{9}{l}{\textit{4. Ablation of sharing decoders layers.}} \\
\midrule
\textbf{E3-8, DC3-8} & 30.1 & 23.9 & 30.6 & 25.2 & 35.0 & 34.5 & 29.9 & 31.5 \\
\textbf{E3-8, C3-8} & 31.4 & 27.6 & 31.7 & 25.7 & 35.4 & 36.9 & \textbf{31.5} & \textbf{31.7} \\
\bottomrule
\end{tabular}
}
\label{tab:interlingua}
\end{table}

The first group in \cref{tab:interlingua} conducts an ablation study of the interlingua layer sharing among the encoders. Obviously, without the interlingua layers, there is a gap of as large as 18.2 points in BLEU score on zero-shot translation when compared with the setup with interlingua layers. This clearly shows that encoder interlingua layers are essential to the zero-shot transfer of translation.

The second group in \cref{tab:interlingua} shows zero-shot translation performance when adjusting the language-specific and interlingua partition of the encoder layers under the condition of fixing other settings. Because the low layers of encoders carry more syntactic information, we treat the top layers as interlingua. The average score of zero-shot translation decreases from 30.1 to 26.2 along with the fewer interlingua layers. This trend is consistent with the observation of \cite{conneau2020emerging} in zero-shot cross-lingual transfer learning using pretrained language model that is equivalent to a Transformer encoder. However, the zero-shot difference between the first two rows (\textbf{E1-8,DC3-8} and \textbf{E3-8,DC3-8}) is as small as 0.2 (30.1-29.9), meaning that certain language-specific layers could be equally important to interlingua layers. As a result, in the main experiment and the incremental language expansion, we choose to share the top 6 layers among the encoders and leave the bottom 3 not shared to intentionally give the model more language-specific capacity while not losing the transfer learning ability in zero-shot translation.

The third group in \cref{tab:interlingua} shows the impact of cross-attention to the zero-shot translation. We compare 3 cases in which the cross-attentions happen in the top, middle, and bottom of the decoder stack, respectively. 
The average zero-shot score increases from 30.9 to 32.1 when cross-attention layers change from top (\textbf{C5-10}) to bottom (\textbf{C1-6}). Intuitively, the middle layers of the decoders capture more language-agnostic information than the top and the bottom. So cross-attentions should only happen in the middle stack. Otherwise, the universal representation from encoders could be messed up by language-specific information. However, the experimental result does not match our intuition. \cite{tan2019study} also points out that the bottom layers of decoders capture more language-agnostic information. One possible explanation could be that comparing with language understanding which most likely happens in the bottom layers, language generation in the top stack requires much more capacity to model the complex language phenomenons. Furthermore, conducting cross-attentions in the bottom layers is also better than in all layers (\textbf{E3-8,C1-10} in the first group), which proves that universal representation should not be messed up by language-specific information.

The last group is an ablation test about whether or not sharing decoder layers. The result shows that not sharing decoder layers achieves higher BLEU scores on both zero-shot (31.5 vs. 29.0) and supervised (31.7 vs. 31.5) translation. Interestingly, when comparing \textbf{C1-10} in group 1 and \textbf{DC3-8} in group 2, and assuming that the impact of different cross-attentions is not more than 2 points (based on the results in group 3), we find that decoder sharing is very beneficial in the case of no encoder sharing. Hence, we suspect that the decoder takes the responsibility to tradeoff between keeping the language characteristics and ensuring the representation universality. As a result, we don't share decoder layers in our experiments. However, for applications that depend on finetuning to transfer the decoder to other language generation tasks, it's worth well to further explore the sharing mechanism of the decoder interlingua layers.

\subsubsection{Conclusion}
Summarised by \cref{tab:interlingua}, the recommended configuration for the proposed interlingua structure is sharing the most top layers of the encoders, not sharing any parameter of the decoders, and conducting cross-attentions in the bottom layers of the decoders. Based on this principle, in our main experiment, we use the configuration \textbf{E4-9,C1-6} for the model with 9 encoder layers and 12 decoder layers.

\subsection{Denoising Auto-encoder (DAE) Task Anaysis}


In this section, 
We conduct an ablation test on the DAE task and compare it with reconstruction task used by previous works \cite{lu2018neural,vazquez2019multilingual}, which we denote as \textbf{AE} (Auto-Encoder) task. 
We use the same experiment settings as in the main experiment unless stated otherwise. The results are shown in \cref{table:dae-analysis}.

\begin{table}[tb]
\caption{Compare the different auxiliary tasks based on zero-shot translation performance. The metric is BLEU. The best scores are in bold.} 
\centering
\footnotesize
\resizebox{1\linewidth}{!}{
\begin{tabular}[b]{l cc cc cc cc}
\toprule
Europarl & \multicolumn{8}{c}{De, Es, Fr $\leftrightarrow$ En} \\
\midrule
\multirow{2}{*}{Direction} & \multicolumn{2}{c}{Fr-De} & \multicolumn{2}{c}{Es-De} & \multicolumn{2}{c}{Es-Fr} & Zero & Parallel \\
& $\leftarrow$ & $\rightarrow$ & $\leftarrow$ & $\rightarrow$ & $\leftarrow$ & $\rightarrow$ & Avg & Avg \\
\midrule
\textbf{MT}              & 0.44 & 0.35 & 0.94 & 0.41 & 0.82 & 0.41 & 0.56 & 33.26 \\
+\textbf{AE}  & 7.31 & 6.25 & 8.32 & 6.39 & 22.67 & 21.79 & 12.12 & 32.73 \\
+\textbf{DAE} & 29.30 & 24.60 & 30.13 & 24.48 & 37.39 & 37.59 & 30.58 & 33.29 \\
+\textbf{DAE}+\textbf{Align} & 29.44 & 24.94 & 30.47 & 25.01 & 38.00 & 37.43 & \textbf{30.88} & \textbf{33.37} \\
\bottomrule
\end{tabular}
}
\label{table:dae-analysis}
\end{table}

When trained only using \textbf{MT} task, the zero-shot BLEU score was less than 1.0 for all zero-shot directions. Similar results are also observed by previous works \cite{lu2018neural,vazquez2019multilingual} using language-specific encoders-decoders. 
To encourage the model to share the encoder representations across English and non-English source sentences, they added an extra identity language pair (\textbf{AE} task) to joint training. The identity pair forces the non-English source embeddings to be close with English source embeddings. We also experiment with \textbf{AE} task. The result is much better than training only with \textbf{MT} task. The average BLEU score on zero-shot translation is 12.12. 
When training with \textbf{DAE} task, we get 30.58 BLEU points on zero-shot translation, which is close to supervised translation (33.29). We conjecture that \textbf{AE} task is too simple for model to learn aligning the non-English sentence representation with English sentence representation. 
To better understand DAE's ability to align the sentence embedding, we also add an explicit alignment loss for the encoder representations used by previous works \cite{arivazhagan2019missing,zhu2020language,escolano2019bilingual} to improve the zero-shot translation. The result (\textbf{DAE+Align}) shows that, comparing \textbf{DAE} task, explicitly aligning the latent representation only increase 0.3 BLEU points on zero-shot translation. It proves that DAE can provide good enough universal representation for zero-shot translation that makes the representation alignment unnecessary.

\section{Conclusion}
In this paper, to improve the zero-shot translation on language-specific encoders-decoders, we first introduce an interlingua to language-specific encoders-decoders via parameter sharing methods on Transformer layers and selective cross-attention between interlingua and decoder. Then we use denoising auto-encoder task to better align the semantic representation of different languages.
Experiments on the Europarl and MultiUN corpora show that our proposed methods significantly improve the zero-shot translation on language-specific encoders-decoders and achieve competitive or better results than universal encoder-decoder counterparts and pivot-based methods while keeping the superiority of adding a new language without retraining the existing modules.

\bibliographystyle{IEEEtran}
\bibliography{IEEEabrv,ijcnn}

\end{document}